\documentclass[10pt,twocolumn,letterpaper]{article}

\usepackage{cvpr}
\usepackage{times}
\usepackage{epsfig}
\usepackage{graphicx}
\usepackage{amsmath}
\usepackage{amssymb}
\usepackage{multirow}

\usepackage[breaklinks=true,bookmarks=false]{hyperref}

\cvprfinalcopy 

\def\cvprPaperID{****} 

\ifcvprfinal\fi
\begin{document}

\title{Learning Rich Image Region Representation for Visual Question Answering}

\author{Bei Liu, Zhicheng Huang, Zhaoyang Zeng, Zheyu Chen, Jianlong Fu\\
Microsoft Research Asia\\
{\tt\small \{bei.liu, t-zhihua, v-zhazen, t-zheche, jianf\}@microsoft.com}
}

\maketitle

\begin{abstract}
We propose to boost VQA by leveraging more powerful feature extractors by improving the representation ability of both visual and text features and the ensemble of models. For visual feature, some detection techniques are used to improve the detector. For text feature, we adopt BERT as the language model and find that it can significantly improve VQA performance. Our solution won the second place in the VQA Challenge 2019.
\end{abstract}

\section{Introduction}

The task of Visual Question Answering (VQA) requires our model to answer text question based on the input image. Most works \cite{Kim2018,peng2018dynamic,yu2018beyond} leverage visual features extracted from images and text features extracted from question to perform classification to obtain answers. Thus, visual and textual features serve as basic components which can directly impact the final performance. In this paper, we propose to improve the performance of VQA by extracting more powerful visual and text features. 

For visual features, most existing works \cite{
Kim2018,peng2018dynamic,yu2018beyond} adopt bottom-up-attention features released by \cite{Anderson2017up-down}, whose feature extractor is a Faster R-CNN object detector built upon a ResNet-101 backbone. We adopt more powerful backbones (i.e. ResNeXt-101, ResNeXt-152) to train stronger detectors. Some techniques (i.e. FPN, multi-scale training) that are useful to improve the accuracy of detectors can also help to boost the performance of VQA.  

For text features, we build upon recent state-of-art techniques in the NLP community. Large-scale language models such as ELMO \cite{Peters:2018}, GPT \cite{radford2018improving} and BERT \cite{devlin2018bert}, have shown excellent results for various NLP tasks in both token and sentence level. BERT uses masked language models to enable pre-trained deep bidirectional representations and allows the representation to fuse the right and left context. While in VQA model, to get the question answer, we need the token level features to contain questions' contextual information to fuse with the visual tokens for the reasoning. So we adopt the BERT as our language mode.

Experiments on VQA 2.0 dataset shows the effectiveness of each component of our solution.  Our final model achieves $74.89\%$ accuracy on \textit{test-standard} split, which won the second place in the VQA Challenge 2019.  

\section{Feature Representation}

\begin{table*}[t]
    \centering
    \begin{tabular}{|c|c|c|c|c|c|c|c|c|}
        \hline
        \textbf{Split} & \textbf{Backbone} & \textbf{FPN dim} & \textbf{Attribute} &\textbf{Language} &\textbf{Yes/No} &\textbf{Num}&\textbf{Others} &\textbf{Score} \\
        \hline
        \multirow{14}{*}{\textit{test-dev}} & Bottom-up-attention (ResNet-101) & - & \checkmark & Glove & 85.42 & 54.04 & 60.52 & 70.04 \\
        & FaceBook pythia (ResNeXt-101) & 512 & \checkmark & Glove & 85.56 & 52.68 & 60.87 & 70.11 \\
        \cline{2-9}
        & ResNeXt-101 & 256 & & Glove & 83.1 & 53.0 & 55.62 & 66.64 \\
        & ResNeXt-101 & 256 & \checkmark & Glove & 85.44 & 54.2 & 60.87 & 70.23 \\
        \cline{2-9}
        & ResNeXt-152 & 256 & \checkmark & Glove & 86.42 & 55.11 & 61.88 & 71.22 \\
        & ResNeXt-152 & 512 & \checkmark & Glove & 86.59 & 56.44 & 62.06 & 71.53 \\
        & ResNeXt-152 (ms-train) & 256 & \checkmark & Glove & 86.46 & 56.37 & 62.24 & 71.55 \\
        & ResNeXt-152 (ms-train) & 512 & \checkmark & Glove & 86.54 & 56.90 & 62.31 & 71.68 \\
        \cline{2-9}
        & ResNeXt-152 & 256 & \checkmark & BERT & 88.00 & 56.28 & 62.90 & 72.48 \\
        & ResNeXt-152 & 512 & \checkmark & BERT & 88.15 & 56.79 & 62.98 & 72.64 \\
        & ResNeXt-152 (ms-train) & 256 & \checkmark & BERT & 88.14 & 56.74 & 63.3 & 72.79 \\
        & ResNeXt-152 (ms-train) & 512 & \checkmark & BERT & 88.18 & 55.35 & 63.16 & 72.58 \\
        \cline{2-9}
        & Ensemble (5 models) & - & - & - & 89.65 & 58.53 & 65.27 & 74.55 \\
        & Ensemble (20 models) & - & - & - & 89.81 & 58.89 & 65.39 & 74.71 \\
        \hline
        \textit{test-std} & Ensemble (20 models) & - & - & - & 89.81 & 58.36 & 65.69 & 74.89 \\
        \hline
    \end{tabular}
    \caption{Experiment results on VQA 2.0 \textit{test-dev} and \textit{test-std} splits. We adopt BAN as VQA model in all settings. The first two rows indicate the results of models we train on released features. ``ms-train'' means using multi-scale strategy in detectors training.}
    \label{tab:my_label}
\vspace{-3mm}
\end{table*}

\subsection{Visual Feature}
Existing works \cite{anderson2018bottom,teney2018tips} show that detection features are more powerful than classification features on VQA task, therefore we train object detectors on large-scale dataset for feature extraction.

\textbf{Dataset.} We adopt Visual Genome 1.2\cite{krishna2017visual} as object detection dataset. Following the setting in \cite{anderson2018bottom}, we adopt $1,600$ object classes and $400$ attribute classes as training categories. The dataset is divided into \textit{train}, \textit{val}, and \textit{test} splits, which contain $98,077$, $5,000$ and $5,000$ images, respectively. We train detectors on \textit{train} split, and use \textit{val} and \textit{test} as validation set to tune parameters.

\textbf{Detector.} We follow the pipeline of Faster R-CNN \cite{ren2015faster} to build our detectors. We adopt ResNeXt \cite{xie2017aggregated} with FPN \cite{lin2017feature} as backbone, and use parameters pretrained on ImageNet \cite{deng2009imagenet} for initialization. We use RoIAlign \cite{he2017mask} to wrap region features into fixed size, and embed them into $2048$-dim via two fully connected layer. Similar to \cite{anderson2018bottom}, we extend a classification branch on the top of region feature, and utilize attribute annotations as additional supervision. Such attribute branch is only used to enhance feature representation ability in the training stage, and will be discarded in the feature extraction stage.

\textbf{Feature.} Given an image, we first feed it into the trained detector, and apply non-maximum suppression (NMS) on each category to remove duplicate bbox. Then we seek $100$ boxes with highest object confidence, and extract their $2048$-dim FC feature. These $100$ boxes with their features are considered as the representation of the given image.

\subsection{Language Feature}
The BERT model, introduced in \cite{devlin2018bert}, can be seen as a multi-layer bidirectional Transformer based on \cite{vaswani2017attention}. The model consists of Embedding layers, Transformer blocks and self-attention heads, which has two different model size. For the base model, there are $12$ layers Transformer blocks, the hidden size is $768$-dim, and the number of self-attention heads is $12$. The total parameters are $110$M. For the large model size, the model consists of $24$ Transformer blocks which hidden size is $1024$, and $16$ self-attention heads. And the model total parameters is $340$M  The model can process a single text sentence or a pair of text sentences (i.e.,[Question, Answer]) in one token sequence. To separate the pair of text sentence, we can add special token ([SEP]) between two sentences, add a learned sentence A embedding to every token of the first sentence and a sentence B embedding to every token of the second sentence. For VQA task, there is only one sentence, we only use the sentence A embedding.

Considering that the total parameter of VQA model is less than $100$M, we use the base BERT as our language model to extract question features. To get each token's representation, we only use the hidden state corresponding to the last attention block features of the full sequence. Pre-trained BERT model has shown to be effective for boosting many natural language processing tasks, we adopt the base BERT uncased pre-train weight as our initial parameters. 

\section{VQA Model}
Recent years, there are many VQA models, which have achieved surprising results. We adopt the Bilinear Attention Networks \cite{Kim2018} (BAN) as our base model. The single model with eight-glimpse can get $70.04$ on VQA2.0 test-dev subset. The BAN model uses Glove and GRU as the language model. And the language feature is a vector $[question length,1280]$. To improve the VQA model performance, we replace the language model with base BERT and modify the BAN language input feature dimension. To Train the BAN with BERT, we use all settings from BAN, but set the max epoch is $20$ with costing learn rate scheduler. To use the base BERT pre-trained parameters we set the learning rate of the BERT module to $5e-5$.

\section{Experiments}

\subsection{Ablation Experiments}
Table~\ref{tab:my_label} shows all our ablation study on each component, including attribute head, FPN dimension, language model. From $3^{rd}$ and $4^{th}$ row, we can find that the attribute head can bring more than $4\%$ improvement to the final performance, which shows the effectiveness of such module. From $5^{th}$ to $12^{th}$ row, we find that BERT can boost the performance by more than $1$ point improvement stably. Besides, increasing FPN dimension and adopting multi-scale training can both slightly improve the VQA accuracy.

\subsection{Comparison with Others}
We select BAN trained on Bottom-up-attention and Facebook features as baselines. Our single model result achieves $72.79\%$ accuracy on \textit{test-dev} split, which significantly outperforms all existing state-of-the-arts. We also ensemble several models we trained by averaging their probabilities output. The result by $20$ models' ensemble achieves $74.71\%$ and $74.89\%$ accuracy on VQA \textit{test-dev} and \textit{test-std} splits, respectively. Such result won the second place in the VQA Challenge 2019.

\section{Conclusion}
We have shown that for VQA task, the representation capacity of both visual and textual features is critical for the final performance.

{\small
\bibliographystyle{ieee_fullname}
\bibliography{egbib}

\begin{thebibliography}{10}\itemsep=-1pt

\bibitem{Anderson2017up-down}
Peter Anderson, Xiaodong He, Chris Buehler, Damien Teney, Mark Johnson, Stephen
  Gould, and Lei Zhang.
\newblock Bottom-up and top-down attention for image captioning and visual
  question answering.
\newblock In {\em CVPR}, 2018.

\bibitem{anderson2018bottom}
Peter Anderson, Xiaodong He, Chris Buehler, Damien Teney, Mark Johnson, Stephen
  Gould, and Lei Zhang.
\newblock Bottom-up and top-down attention for image captioning and visual
  question answering.
\newblock In {\em Proceedings of the IEEE Conference on Computer Vision and
  Pattern Recognition}, pages 6077--6086, 2018.

\bibitem{deng2009imagenet}
Jia Deng, Wei Dong, Richard Socher, Li-Jia Li, Kai Li, and Li Fei-Fei.
\newblock Imagenet: A large-scale hierarchical image database.
\newblock In {\em 2009 IEEE conference on computer vision and pattern
  recognition}, pages 248--255. Ieee, 2009.

\bibitem{devlin2018bert}
Jacob Devlin, Ming-Wei Chang, Kenton Lee, and Kristina Toutanova.
\newblock Bert: Pre-training of deep bidirectional transformers for language
  understanding.
\newblock {\em arXiv preprint arXiv:1810.04805}, 2018.

\bibitem{he2017mask}
Kaiming He, Georgia Gkioxari, Piotr Doll{\'a}r, and Ross Girshick.
\newblock Mask r-cnn.
\newblock In {\em Proceedings of the IEEE international conference on computer
  vision}, pages 2961--2969, 2017.

\bibitem{Kim2018}
Jin-Hwa Kim, Jaehyun Jun, and Byoung-Tak Zhang.
\newblock {Bilinear Attention Networks}.
\newblock In {\em Advances in Neural Information Processing Systems 31}, pages
  1571--1581, 2018.

\bibitem{krishna2017visual}
Ranjay Krishna, Yuke Zhu, Oliver Groth, Justin Johnson, Kenji Hata, Joshua
  Kravitz, Stephanie Chen, Yannis Kalantidis, Li-Jia Li, David~A Shamma, et~al.
\newblock Visual genome: Connecting language and vision using crowdsourced
  dense image annotations.
\newblock {\em International Journal of Computer Vision}, 123(1):32--73, 2017.

\bibitem{lin2017feature}
Tsung-Yi Lin, Piotr Doll{\'a}r, Ross Girshick, Kaiming He, Bharath Hariharan,
  and Serge Belongie.
\newblock Feature pyramid networks for object detection.
\newblock In {\em Proceedings of the IEEE Conference on Computer Vision and
  Pattern Recognition}, pages 2117--2125, 2017.

\bibitem{peng2018dynamic}
Gao Peng, Hongsheng Li, Haoxuan You, Zhengkai Jiang, Pan Lu, Steven Hoi, and
  Xiaogang Wang.
\newblock Dynamic fusion with intra-and inter-modality attention flow for
  visual question answering.
\newblock {\em arXiv preprint arXiv:1812.05252}, 2018.

\bibitem{Peters:2018}
Matthew~E. Peters, Mark Neumann, Mohit Iyyer, Matt Gardner, Christopher Clark,
  Kenton Lee, and Luke Zettlemoyer.
\newblock Deep contextualized word representations.
\newblock In {\em Proc. of NAACL}, 2018.

\bibitem{radford2018improving}
Alec Radford, Karthik Narasimhan, Time Salimans, and Ilya Sutskever.
\newblock Improving language understanding with unsupervised learning.
\newblock Technical report, Technical report, OpenAI, 2018.

\bibitem{ren2015faster}
Shaoqing Ren, Kaiming He, Ross Girshick, and Jian Sun.
\newblock Faster r-cnn: Towards real-time object detection with region proposal
  networks.
\newblock In {\em Advances in neural information processing systems}, pages
  91--99, 2015.

\bibitem{teney2018tips}
Damien Teney, Peter Anderson, Xiaodong He, and Anton van~den Hengel.
\newblock Tips and tricks for visual question answering: Learnings from the
  2017 challenge.
\newblock In {\em Proceedings of the IEEE Conference on Computer Vision and
  Pattern Recognition}, pages 4223--4232, 2018.

\bibitem{vaswani2017attention}
Ashish Vaswani, Noam Shazeer, Niki Parmar, Jakob Uszkoreit, Llion Jones,
  Aidan~N Gomez, {\L}ukasz Kaiser, and Illia Polosukhin.
\newblock Attention is all you need.
\newblock In {\em Advances in neural information processing systems}, pages
  5998--6008, 2017.

\bibitem{xie2017aggregated}
Saining Xie, Ross Girshick, Piotr Doll{\'a}r, Zhuowen Tu, and Kaiming He.
\newblock Aggregated residual transformations for deep neural networks.
\newblock In {\em Proceedings of the IEEE conference on computer vision and
  pattern recognition}, pages 1492--1500, 2017.

\bibitem{yu2018beyond}
Zhou Yu, Jun Yu, Chenchao Xiang, Jianping Fan, and Dacheng Tao.
\newblock Beyond bilinear: Generalized multimodal factorized high-order pooling
  for visual question answering.
\newblock {\em IEEE Transactions on Neural Networks and Learning Systems},
  29(12):5947--5959, 2018.

\end{thebibliography}
}

\end{document}